\documentclass[dvipsnames,format=sigconf,anonymous=false,review=false]{acmart}
\AtBeginDocument{%
  }

\copyrightyear{2025}
\acmYear{2025}
\setcopyright{rightsretained}
\acmConference[{\upshape This is the author's version of the work. It is posted here for your personal use. Not for redistribution. The definitive Version of Record was published in} GECCO '25]{Genetic and Evolutionary Computation Conference}{July 14--18, 2025}{Malaga, Spain}
\acmBooktitle{Genetic and Evolutionary Computation Conference (GECCO '25), July 14--18, 2025, Malaga, Spain}
\acmDOI{10.1145/3712256.3726370}
\acmISBN{979-8-4007-1465-8/2025/07}

\usepackage{graphicx} 
\usepackage{array}
\usepackage{textcomp}
\usepackage{stfloats}
\usepackage{url}
\usepackage{verbatim}
\usepackage{wrapfig}
\usepackage{subcaption}
\usepackage{graphicx}
\usepackage{booktabs}
\usepackage{algorithm}
\usepackage{algorithmic}
\usepackage{amsmath}
\usepackage{amsfonts}
\usepackage{mathtools}
\usepackage{stmaryrd}
\usepackage{bm}
\usepackage{bbm}
\usepackage{amsthm}
\usepackage{chngpage}
 
\providecommand{\argmin}{\operatornamewithlimits{argmin}}

\providecommand{\R}{\mathbb{R}} 
\providecommand{\E}{\mathbb{E}} 
\providecommand{\T}{\mathrm{T}}

\DeclarePairedDelimiterX{\inner}[2]{\langle}{\rangle}{#1, #2}
\DeclarePairedDelimiter{\norm}{\lVert}{\rVert}
\DeclarePairedDelimiter{\abs}{\lvert}{\rvert}

\usepackage{cleveref}

\theoremstyle{definition}

\usepackage{ifthen}
\usepackage{tabularx}
\usepackage{multirow}
\newcommand{\markupdraft}[2]{
\ifthenelse{\equal{#1}{display}}{#2}{}
\ifthenelse{\equal{#1}{color}}{\color{#2}}{}
}

\newcommand{\newcolored}[3][]{{\markupdraft{color}{#2}#3}
\ifthenelse{\equal{#1}{}}{}{\markupdraft{display}{{\color{yellow!70! Black}[#1]}}}}
\newcommand{\del}[2][]{{\markupdraft{display}{{\color{orange}[removed: ``#2''[#1]]}}}}

\renewcommand{\del}[2]{}

\begin{document}

\title{Challenges of Interaction in Optimizing Mixed Categorical-Continuous Variables}

\author{Youhei Akimoto}
\authornote{All authors contributed equally to this research.}
\email{akimoto@cs.tsukuba.ac.jp }
\orcid{0000-0003-2760-8123}
\affiliation{%
  \institution{University of Tsukuba \& RIKEN AIP}
  \city{Tsukuba}
  \state{Ibaraki}
  \country{JAPAN}
}
\author{Xilin Gao}
\authornotemark[1]
\email{xilin.gao@bbo.cs.tsukuba.ac.jp}
\orcid{0009-0005-8218-6375}
\affiliation{%
  \institution{University of Tsukuba}
  \city{Tsukuba}
  \state{Ibaraki}
  \country{JAPAN}
}
\author{Ze Kai Ng}
\authornotemark[1]
\email{ngzekai@bbo.cs.tsukuba.ac.jp}
\orcid{0009-0008-6114-0388}
\affiliation{%
  \institution{University of Tsukuba}
  \city{Tsukuba}
  \state{Ibaraki}
  \country{JAPAN}
}
\author{Daiki Morinaga}
\authornotemark[1]
\email{ morinaga@bbo.cs.tsukuba.ac.jp}
\orcid{0009-0009-7937-7567}
\affiliation{%
  \institution{University of Tsukuba \& RIKEN AIP}
  \city{Tsukuba}
  \state{Ibaraki}
  \country{JAPAN}
}


\begin{abstract}
Optimization of mixed categorical-continuous variables is prevalent in real-world applications of black-box optimization. Recently, CatCMA has been proposed as a method for optimizing such variables and has demonstrated success in hyper-parameter optimization problems. However, it encounters challenges when optimizing categorical variables in the presence of interaction between continuous and categorical variables in the objective function. In this paper, we focus on optimizing mixed binary-continuous variables as a special case and identify two types of variable interactions that make the problem particularly challenging for CatCMA. To address these difficulties, we propose two algorithmic components: a warm-starting strategy and a hyper-representation technique. We analyze their theoretical impact on test problems exhibiting these interaction properties. Empirical results demonstrate that the proposed components effectively address the identified challenges, and CatCMA enhanced with these components, named ICatCMA, outperforms the original CatCMA.
\end{abstract}



\maketitle


\section{Introduction}
In black-box optimization (BBO), it is common to encounter domains of design variables that consist of both discrete and continuous elements, such as integer-continuous and categorical-continuous variables~\cite{hazan2018hyperparameter, akiba2024evolutionary,lu2020mixed}.
Problems involving integer and continuous variables are called Mixed-Integer Optimization, while those involving categorical and continuous variables are called Mixed-Categorical Optimization. Collectively, these are known as Mixed-Variable Optimization.
The most significant difference between integer variables and categorical variables is that, unlike integer ones, categorical ones do not have an ordinal relationship between values.

Several methods have been proposed to address such distinctive challenges of mixed categorical-continuous problems.
In the context of Bayesian optimization, Ru et al.~\cite{ru2020bayesian} first proposed a method for mixed categorical-continuous problems, which
uses the multi-armed bandits approach for categorical variables and Gaussian process-based Bayesian optimization approach for continuous variables and connects them with a tailored kernel.
Another Bayesian optimization approach~\cite{wan2021think} is proposed which is more robust to high-dimensional problems and is theoretically guaranteed to converge to the global optimum.
While Bayesian optimization approaches (see~\cite{wan2021think} for other works) effectively handle the challenges of mixed categorical-continuous problems, they are hindered by scalability issues related to dimensionality and the number of evaluations, due to the high computational costs associated with Gaussian processes.

A pioneering work by Hamano et al.~\cite{hamano2024catcma} proposes CatCMA, a variant of the well-known black-box continuous optimization, namely, covariance matrix adaptation evolution strategy (CMA-ES)~\cite{hansen2001completely, hansen2003reducing, hansen2006cma}, which is known to be scalable~\cite{santoni2024comparison}.
In this study, the authors used a joint probability distribution to simultaneously sample categorical and continuous variables. They update this model using a rule derived from information geometric optimization (IGO)~\cite{ollivier2017igo}, which is an extended optimization framework of CMA-ES.
It is demonstrated that CatCMA is more scalable to high-dimensional problems than existing mixed categorical-continuous optimization algorithms.

It is reasonable to assume that the interaction between categorical and continuous variables contributes significantly to the complexity of the optimization problem.
In~\cite{hamano2024catcma}, CatCMA has been shown to effectively solve a problem where categorical variables weakly influence the importance of continuous variables.
In this study, the more challenging interactions are studied (more formally introduced in \Cref{sec:prob}).
The first interaction type (type-I) refers to cases where whether continuous variables contribute to the objective function depends on the categorical variables.
Specifically, under certain categorical variables, some dimensions of the continuous variables are masked so that only selected continuous variables affect the objective function value. Such an interaction appears in~\cite{akiba2024evolutionary} for instance. 
The second interaction type (type-II) refers to cases where the values of the continuous variables that minimize the objective function depend on the categorical variables. This type of interaction is very intuitive and naturally appears in real-world problems, yet it remains to be investigated.

In this paper, we propose methods to handle type-I interaction and type-II interaction in the mixed categorical-continuous optimization problem.
For the sake of simplicity, we limit our attention to the mixed binary-continuous variables, a special case of the mixed categorical-continuous variables. 
Our first contribution is that we empirically reveal that the CatCMA fails to optimize test problems with type-I interaction and type-II interaction.
Our second contribution is that we propose two simple versatile modifications of the optimization algorithm and empirically demonstrate their effectiveness by applying them to the CatCMA.
The first proposed component, warm-starting, freezes the update of the search distribution for the binary variables at the beginning while updating the distribution for the continuous variables. 
With warm-starting, the optimizer prevents the risk of premature convergence of the categorical variables due to suboptimal continuous variables.
The second proposed component, the hyper-representation, is designed to effectively reduce type-II interaction between the binary and continuous variables by introducing a parameterized map that takes the binary variables as its input and outputs the corresponding continuous variables. 
The optimizer then attempts to solve the modified mixed binary-continuous problem, where the continuous variables are replaced with the parameters of the map.

This paper is organized as follows.
We describe the formulation of mixed binary-continuous problems and introduce two aforementioned interaction types in \Cref{sec:prob}.
\Cref{sec:catcma} is devoted to describe CatCMA.
We introduce the warm-starting strategy and the hyper-representation technique in \Cref{sec:warm-start} and \Cref{sec:hyper-representation}, respectively. 
Experiments on test problems in \Cref{sec:experiments} reveal the effect of each proposed component and show both components improve the performance of CatCMA on problems with interaction types I and II. 
We summarize this work and discuss the limitation of the current study in \Cref{sec:conclusion}.

\section{Mixed Binary-Continuous Problems}\label{sec:prob}

Optimization of a function $f: \mathcal{C} \times \mathcal{X} \to \R$ that takes a binary vector $c \in \mathcal{C} = \{0, 1\}^{m}$ and a continuous vector $x \in \mathcal{X} = \R^{n}$ is called a mixed binary-continuous problem. 
A simple and easy-to-solve example problem is a separable problem where the objective function can be decomposed as
\begin{equation}
f(c, x) = f_c(c) + f_x(x).
\end{equation}
In this case, optimization can be solved for $c$ while fixing $x$, then solved for $x$ while fixing $c$, or vice versa.
We are interested in more practical situations where binary and continuous vectors interact. 

\paragraph{Type-I interaction}
The first type of interaction is the dependency of effective continuous dimensions on a binary vector. 
A simple problem with the type-I interaction is constructed as follows.
Suppose $n = m$, that is, the dimensions of the binary vectors and the continuous vectors are the same.
Let $\odot$ denote the Hadamard product (i.e., element-wise product). 
Let $b^\star \in \R^n$ be a predefined vector whose elements must be nonzero. 
Then, we define
\begin{equation}
    f_{I}(c, x) = \norm{ x \odot c - b^\star}^2 .
    \label{eq:f_I}
\end{equation}
The unique optimal solution\footnote{It is necessary for the uniqueness of the optimal solution that all elements of $b^\star$ are nonzero.} is a pair consisting of $c^\star = (1, \dots, 1)$ and $x^\star = b^\star$. 
The optimal continuous vector $x^\star(c)$ under given $c$ is 
\begin{equation}
[x^\star(c)]_{i} = 
\begin{cases}
[b^\star]_{i}  & [c]_i = 1,  \\
\text{arbitrary}  & [c]_i = 0,  
\end{cases}\label{eq:f1x}
\end{equation}
where $[\cdot]_i$ represents the $i$-th coordinate of a vector.
Though it is not uniquely determined, $b^\star$ is always the optimal vector independently of $c$. 
In this sense, this problem does not have the difficulty of the type-II interaction discussed below. 
However, the optimal binary variables change depending on the continuous variables.
The optimal binary variables are
\begin{equation}
[c^\star(x)]_i = 
\begin{cases}
1 & \abs{[x - b^\star]_i} \leq \abs{[b^\star]_i},  \\
0 & \text{otherwise}.  
\end{cases}
\end{equation}
Unless the continuous variables are sufficiently close to their optimal values, the optimal binary variables flip.

\paragraph{Type-II interaction} 
The second type of interaction we focus on is the dependence of the optimal value for the continuous variables on the choice of the binary variables. Formally, it is stated as follows.
Let $\phi^\star: \mathcal{C} \to \mathcal{X}$ be the map defining the optimal continuous values:
\begin{equation}
\phi^\star(c) := \argmin_{x \in \mathcal{X}} f(c, x).\label{eq:phistar}
\end{equation}
For the sake of simplicity, we assume the uniqueness of the optimal value $\phi^\star(c)$ for each $c \in \mathcal{C}$ at the moment.
In the above separable problem, it is easy to see that $\phi^\star$ is constant over $\mathcal{C}$. 
However, in general, $\phi^\star$ changes over $\mathcal{C}$ and makes the problem difficult. 
For example, the aforementioned alternating approach may not work.

A simple example problem with type-II interaction can be constructed below. 
Let $\phi^\star$ be predefined. For example, let $V^\star \in \R^{n \times m}$ and $b^\star \in \R^n$ and define $\phi^\star$ as $\phi^\star(c) = V^\star c + b^\star$. 
Let $f_c: \mathcal{C} \to \R$ be a function depending only on $c$. 
For example, $f_c(c) = \sum_{i=1}^{m} (1 - [c]_i)$. 
Then, we define
\begin{equation}
f_{II}(c, x) = f_c(c) + \norm{x - \phi^\star(c)}^2 .\label{eq:f_II}
\end{equation}
The optimal solution is a pair consisting of $c^\star = \argmin_{c \in \mathcal{C}} f_c(c)$ and $x^\star = \phi^\star(c^\star)$.
The optimal continuous vector changes as the binary vector changes. 
Moreover, the optimal binary vector may depend on the continuous vector as the optimal binary vector may change for the second term. 

In this synthetic problem, the strength of the interaction can be controlled by replacing $V^\star$ with $\alpha V^\star$, where $\alpha \geq 0$ is a scalar factor controlling the strength and $V^\star$ is normalized. 
If $\alpha = 0$, the problem reduces to a separable problem $f(c, x) = f_c(c) + \norm{x - b^\star}^2$. 
If $\alpha > 0$ is sufficiently small, the optimal $c$ is determined solely by $f_c$. 
If $\alpha$ is sufficiently large, strong interaction will appear.

\section{C\MakeLowercase{at}CMA}\label{sec:catcma}
CatCMA is a randomized search method for mixed categorical-continuous black-box optimization problems. 
It combines the well-known CMA-ES \cite{hansen2001completely,hansen2003reducing,hansen2006cma} for black-box continuous optimization with the population-based incremental learning algorithm \cite{BALUJA199538} for black-box binary optimization from the viewpoint of information-geometric optimization framework \cite{ollivier2017igo}. 
In this section, we introduce CatCMA for mixed binary-continuous optimization problems.

CatCMA repeats the generation of candidate solutions from a sampling distribution, the evaluation of candidate solutions on the objective function, and the update of the distribution parameters.
Each candidate solution is a pair of binary variables and continuous variables. 
The sampling distribution is modeled as the joint distribution of the Bernoulli distribution and the Gaussian distribution.
The probability mass function of the Bernoulli distribution is defined as
\begin{equation}
p(c \mid q) = \prod_{i=1}^m [q]_{i}^{[c]_{i}} (1 - [q]_i)^{1 - [c]_{i}} \enspace, 
\end{equation}
where $[c]_i$ represents the $i$-th element of $c$ and the distribution parameter $q \in [0, 1]^{m}$ consists of the probability $[q]_i$ of $[c]_i$ being $1$.
The probability density function of the Gaussian distribution is defined as
\begin{equation}
p(x \mid \mu, \Sigma) = ((2 \pi)^n \det(\Sigma))^{-\frac{1}{2}} \exp\left( - \frac{ (x - \mu)^\T \Sigma^{-1} (x - \mu) }{2}\right), 
\end{equation}
where $\mu \in \R^n$ and $\Sigma \in \R^{n \times n}$ are the mean vector and the covariance matrix, respectively.
The joint distribution is written as
\begin{equation}
    p(c, x \mid q, \mu, \Sigma) = p(c \mid q)  p(x \mid \mu, \Sigma)\enspace.
\end{equation}

The update of the distribution parameters mainly follows the IGO framework. 
Roughly speaking, the distribution parameters, $\theta = (q, \mu, \Sigma)$, are updated by taking the natural gradient step \cite{amari1998ng} of the expected objective function 
\begin{equation}
J(\theta) = \E_{(c, x) \sim p(\cdot \mid \theta)}[ f(c, x) ] \enspace.
\end{equation}
The natural gradient can be estimated by using $\lambda$ candidate solutions $\{(c_k, x_k)\}_{k=1}^{\lambda}$ independently sampled from the current distribution $p(\cdot \mid \theta^{(t)})$ as
\begin{equation}
    \widehat{\tilde{\nabla}_{\theta} J(\theta^{(t)})} = \frac{1}{\lambda}\sum_{k=1}^\lambda f(c_k, x_k) \tilde{\nabla}_\theta \ln p(c_k, x_{k} \mid \theta^{(t)} ) \enspace.
\end{equation}
The IGO algorithms replace the function values $f(c_k, x_k)$ to the ranking-based weight values $\lambda w_{\mathrm{rk}(k)}$, where $\mathrm{rk}(k)$ is the ranking of $k$-th candidate solutions $(c_k, x_k)$ among $\lambda$ and $\{w_k\}_{k=1}^\lambda$ are the predefined weights, resulting in the parameter update invariant to the monotonic transformations of the objective function.

In practice, the update of the parameters of the Gaussian distribution follows CMA-ES and the update of the parameters of the Bernoulli distribution follows ASNG \cite{akimoto2019adaptive}. Both are regarded as the IGO framework for continuous and binary search domains, but they incorporate several mechanisms that improve the robustness and the efficiency of their search significantly. 
\Cref{alg:catcma} summarizes the algorithm of CatCMA.
The detailed update is as follows.
The mean vector $\mu$ of the Gaussian distribution and the probability vector $q$ of the Bernoulli distribution are updated with the natural gradient estimate with the ranking-based weights as
\begin{align}
    \mu^{(t+1)} &= \mu^{(t)} +  c_m \sum_{k=1}^{\lambda} w^+_{\mathrm{rk}(k)}\left(x_{k} - \mu^{(t)}\right) \label{eq:mean_updatee} \\
    q^{(t+1)} &= q^{(t)} + \eta \sum_{k=1}^\lambda w_{\mathrm{rk}(k)} \left(c_{k} - q^{(t)} \right)\enspace, \label{eq:q_update}
\end{align}
where $c_m$ and $\eta$ are the learning rates and $w_k^+ = \max(w_k, 0)$. Note that we set $w_k \geq 0$, hence $w_k^+ = w_k$ for all $k$.

The update of the covariance matrix is accelerated by incorporating the cumulative step-size adaptation and the rank-one update from CMA-ES~\cite{ostermeier1994csa,hansen1996cma,hansen2001completely}.
For this purpose, the step size $\sigma \in \R_{>0}$ is introduced and the covariance matrix of the Gaussian distribution is parameterized as $\Sigma = \sigma^2 C$.
For the cumulative step-size adaptation and for the rank-one covariance matrix update, 
we maintain evolution paths that accumulate the previous update directions of the distribution mean as
\begin{align}
    p_\sigma^{(t+1)} &= (1 - c_\sigma)p_\sigma^{(t)} + \sqrt{c_\sigma(2-c_\sigma)} \frac{ {C^{(t)}}^{-\frac{1}{2}}(\mu^{(t+1)} - \mu^{(t)}) }{ c_m \sqrt{\sum_{\ell = 1}^{\lambda} (w_\ell^+)^2}} \enspace, \label{eq:sigma_evol_path}
    \\
    p_c^{(t+1)} &= (1 - c_c)p_c^{(t)} + h_\sigma\sqrt{c_c(2-c_c)} \frac{{\sigma^{(t)}}^{-1} (\mu^{(t+1)} - \mu^{(t)}) }{ c_m \sqrt{\sum_{\ell = 1}^{\lambda} (w_\ell^+)^2}} \enspace, \label{eq:cov_evol_path}
\end{align}
where $c_\sigma$  and $c_c$ are the cumulation factors, and $h_\sigma$ is introduced to avoid $p_c$ from being too long when $\sigma$ is too small and $h_\sigma = 1$ if 
\begin{equation}
    \|p_\sigma^{(t+1)}\| < \sqrt{1 - (1 - c_\sigma)^{2(t+1)}}\left(1.4 + \frac{2}{n+1}\right)\E[\|\mathcal{N}(0, I)\|] ,
\end{equation}
otherwise $h_\sigma = 0$. 
The step-size is updated as
\begin{equation}
    \sigma^{(t+1)} = \sigma^{(t)} \exp\left(\frac{c_\sigma}{d_\sigma}\left(\frac{\|p_\sigma^{(t+1)}\|}{\E[\|\mathcal{N}(0, I)\|]}\right)-1\right)\enspace, \label{eq:step_size_update}
\end{equation}
where $d_\sigma$ is the damping factor.
The covariance matrix $C^{(t)}$ is updated as
\begin{multline}\label{eq:cov_update}
    C^{(t+1)} = C^{(t)} +  c_\mu\sum_{k=1}^\lambda w_{\mathrm{rk}(k)}\left(\operatorname{OP}\left(\frac{x_{k} - \mu^{(t)}}{\sigma^{(t)}}\right) - C^{(t)}\right) \\
    + c_1\left(\operatorname{OP}(p_c^{(t+1)}) - (1 - (1 - h_\sigma)c_c(2-c_c))C^{(t)}\right) \enspace, 
\end{multline}
where $c_1$ and $c_\mu$ are the learning rates for rank-one update and rank-$\mu$ update, respectively, and $\operatorname{OP}(x) = x x^\T$ is the self outer product operator.
To avoid numerical errors in the eigenvalue decomposition,  a post-process for step-size $\sigma$ is implemented as
\begin{equation}
    \sigma^{(t+1)} \leftarrow \max\left\{\sigma^{(t+1_)}, \sqrt{\frac{\Lambda^{\min}}{\min\{\text{eig}(C^{(t+1)})\}}}\right\} \enspace, \label{eq:post_process}
\end{equation}
where $\text{eig}(C^{(t+1)})$ is the set of the eigenvalues of $C^{(t+1)}$, and $\Lambda^{\min}$ is the lower bound of the eigenvalues which is set to $\Lambda^{\min} = 10^{-30}$.

To mitigate unbalanced updates between the multivariate Gaussian and categorical distributions, which arise due to the sensitivity to hyperparameter settings such as learning rate, CatCMA employs learning rate adaptation from ASNG into the update process of the categorical distribution, in our case, which is the Bernoulli distribution. 
Let $G^{(t)} = \sum_{k=1}^\lambda w_{\mathrm{rk}(k)} \left(c_{k} - q^{(t)} \right)$.
The learning rate $\eta$ in \eqref{eq:q_update} is computed as
\begin{equation}
\eta^{(t)} = \frac{\delta^{(t)}}{ \norm{G^{(t)}}_{F(q^{(t)})} },\label{eq:eta}
\end{equation}
where $\norm{x}_{M} = \sqrt{x^\T M x}$ is the Mahalanobis norm with respect to the matrix $M$ and $F(q^{(t)})$ is the Fisher information matrix of the Bernoulli distribution, which is a diagonal matrix whose $i$-th diagonal element is $[F(q^{(t)})]_{i,i} = ([q]_i^{(t)} (1 - [q]_i^{(t)}))^{-1}$. 
The denominator in \eqref{eq:eta} plays as a gradient normalization.
The numerator $\delta^{(t)}$ is adapted in a way similar to the cumulative step-size adaptation as follows.
First, the accumulation of the estimated natural gradient is computed as
\begin{align}
    s^{(t+1)} &= (1-\beta)s^{(t)} + \sqrt{\beta(2 - \beta)}F^{\frac{1}{2}}(q^{(t)})G^{(t)} \enspace, \label{eq:s_update}\\
    \gamma^{(t+1)} &= (1 - \beta)^2 \gamma^{(t)} + \beta(2-\beta) \norm{G^{(t)}}^2_{F(q^{(t)})}\enspace, \label{eq:gamma_update}
\end{align}
where $\beta = \delta^{(t)} / m^\frac{1}{2}$. Setting a constant $\alpha=1.5$, $\delta^{(t)}$ is updated as
\begin{equation}
    \delta^{(t+1)} = \delta^{(t)}\exp\left(\beta\left(\frac{\|s^{(t+1)}\|^2}{\alpha} - \gamma^{(t+1)}\right)\right) \enspace. \label{eq:trust_region_update}
\end{equation}

To prevent probability parameters from being $0$ or $1$, where all candidate solutions have the same binary values and these probability parameters never move, the margins are introduced for the probability parameters.
Let $q^\text{min} \in (0, 1)^{m}$ and $q^\text{max} \in (0, 1)^m$ be the vectors representing the lower and upper bounds of the probability vector respectively and defined as
\begin{equation}
    [q^{\text{min}}]_i = 1 - (1 - \xi)^{\frac{1}{m}}
\end{equation}
and $q^\text{max} = 1 - q^\text{min}$, where $\xi=0.27$. 
Then, the probability vector is simply clipped as
\begin{equation}
    [q^{(t+1)}]_{i} \leftarrow 
    \begin{cases}
    [q^{\min}]_i & \text{if } [q^{(t+1)}]_{i} < [q^{\min}]_i; \\
    [q^{(t+1)}]_{i} & \text{if } [q^{\min}]_i \leq [q^{(t+1)}]_{i} \leq [q^{\max}]_i; \\
    [q^{\max}]_{i} & \text{if } [q^{(t+1)}]_{i} > [q^{\min}]_i \enspace.
    \end{cases}
    \label{eq:max_correction} 
\end{equation}

\begin{algorithm}[t]
\caption{CatCMA} \label{alg:catcma}
\begin{algorithmic}[1]
\REQUIRE $f:\mathcal{C}\times\mathcal{X}\to\R$, $\mu^{(0)}\in\mathcal{X}$, $\sigma^{(0)}\in\R_{>0}$, $C^{(0)}\in\R^{n\times n}$, $q^{(0)}\in[0, 1]^m$
\STATE $p_\sigma^{(0)}=p_c^{(0)}=0, \delta^{(0)}=1, s^{(0)}=0, \gamma^{(0)}=0, t=0$
\WHILE{termination conditions are not met}
    \FOR{$k=1, \dots, \lambda$}
        \STATE \textit{Sample} $x_k \sim p\left(x\Bigm|\mu^{(t)}, \left({\sigma^{(t)}}\right)^2 C^{(t)}\right)$
        \STATE \textit{Sample} $c_k \sim p\left(c\bigm|q^{(t)}\right)$
    \ENDFOR
    \STATE \textit{Query} $\{f(c_k, x_k)\}_{k=1}^\lambda$
    \STATE \textit{Update} $\mu^{(t)}$ \textit{using} \eqref{eq:mean_updatee}
    \STATE \textit{Update} $p_\sigma^{(t)}$ \textit{and} $p_c^{(t)}$ \textit{using} \eqref{eq:sigma_evol_path} \textit{and} \eqref{eq:cov_evol_path}
    \STATE \textit{Update} $\sigma^{(t)}$ \textit{and} $C^{(t)}$ \textit{using} \eqref{eq:step_size_update} \textit{and} \eqref{eq:cov_update}
    \STATE Modify $\sigma^{(t+1)}$ \textit{using} \eqref{eq:post_process}
    \STATE \textit{Update} $q^{(t)}$ \textit{using} \eqref{eq:q_update} with $\eta^{(t)}$ computed in \eqref{eq:eta}
    \STATE \textit{Update} $s^{(t)}$ \textit{and} $\gamma^{(t)}$ \textit{using} \eqref{eq:s_update} \textit{and} \eqref{eq:gamma_update}
    \STATE \textit{Update} $\delta^{(t)}$ \textit{using} \eqref{eq:trust_region_update}
    \STATE \textit{Modify} $q^{(t+1)}$ \textit{using} \eqref{eq:max_correction}
    \STATE $t \leftarrow t +1$
\ENDWHILE
\end{algorithmic}
\end{algorithm}

In CatCMA, the joint probability distribution is constructed with two independent probability distribution models, implying that the sampling distribution cannot model the interaction between numerical and categorical variables. Despite CatCMA being empirically shown to solve a test problem that is similar to $f_{II}$ with $\alpha=1$, it has weak interaction between the categorical and continuous variables~\cite{hamano2024catcma}.
Problems such as interaction type-I and type-II that contain strong dependencies between numerical and categorical (binary) variables are difficult to solve with only such joint probability distribution as demonstrated in the experiment section. Consequently, a sophisticated strategy is required to effectively capture the complex interactions between numerical and categorical variables, enabling improved performance in solving such challenging problems.

\section{Warm-Starting Strategy}\label{sec:warm-start}

We address the difficulty of type-I interaction by freezing the update of the probability vector $q$ while optimizing the continuous vector at the beginning of the search. 
In other words, we initialize the search distribution for the continuous vectors by running CMA-ES (i.e., only the continuous part of CatCMA) for some iterations before starting the optimization process of CatCMA. 
From this perspective, we call it the \emph{warm-starting} strategy.
Let $T_{\mathrm{freeze}}$ be the number of iterations we stall the update of $q$. 
By doing so, we expect that $x$ (and its search distribution) is optimized for the objective function expected over the Bernoulli distribution with probability vector $q$, namely,
\begin{equation}
J(x) = \E_{c \sim p(c\mid q)}\left[ f(c, x) \right]. 
\end{equation}

However, simply stalling the update of $q$ is insufficient to optimize $J(x)$. 
The reason is as follows. 
In CatCMA, a pair $(c, x)$ is treated as a single solution to the problem.
Multiple candidate solutions $\{(c_{k}, x_{k})\}_{k=1}^\lambda$ generated at each iteration have different binary vectors, i.e., $c_{k} \neq c_{\ell}$, while they follow the same distribution with parameter $q$. 
Then, the solutions are evaluated on the objective function $f(c_{k}, x_{k})$ and their rankings are computed based on $\{f(c_{k}, x_{k})\}_{k=1}^{\lambda}$. 
The point is that even though we want to update only the distribution for the continuous variables, $x$, for which $\{c_{k}\}_{k=1}^\lambda$ are not used, the ranking of $\{x_{k}\}_{k=1}^{\lambda}$ depends on them. 
In other words, we use the ranking of the noise-corrupted value $f(c_{k}, x_{k})$ instead of that of $J(x_{k})$. 
Therefore, to optimize $J(x)$, a noise handling is required.

As a simple approach, we use the same binary vector for $\lambda$ candidates at each iteration.
At each of the first $T_\mathrm{freeze}$ iterations, we sample a binary vector $c$ from the Bernoulli distribution with $q$, and use it for all the $\lambda$ candidates $\{x_{k}\}_{k=1}^{\lambda}$. 
Then, the candidate solutions are evaluated by $f(c, x_{k})$ with the same $c$. 
Though $c$ itself is randomly generated, the continuous variables are deterministically ranked for each $c$.

\paragraph{Why does it help?}
Let us consider the ideal situation, where there is the optimal continuous vector, $x^\star$, that is optimal for all $c$ just like $f_I$, that is,
\begin{equation}
x^\star \in \argmin_{x \in \mathcal{X}} f(c, x), \quad \forall c \in \mathcal{C}.
\end{equation}
In this case, it is easy to see that $x^\star \in \argmin_{x \in \mathcal{X}} J(x)$ and any solution minimizing $J(x)$ minimizes $f(c, x)$ for all $c$ as long as the distribution of $c$ has a positive probability mass for all $c \in \mathcal{C}$. 
Therefore, even though the objective function $f$ is not separable, solving $J(x)$ first and then solving $f(c, x) \approx f(c, x^\star)$ is expected to result in the optimal solution $(c^\star, x^\star)$. 

Without the warm-starting strategy, the algorithm often fails to locate the optimal solution on $f_I$ for the following reasons. 
If we change the distribution of $c$ from the beginning, the distribution of $c$ has a risk of drift to a boundary of the distribution space where the probability mass is concentrated to a single $c^\dagger$. 
In this situation, because of \eqref{eq:f1x}, there is a high risk of converging to $x^\dagger$ that is optimal for $c^\dagger$ but is different from $x^\star$. 
Once the search distribution for $x$ is concentrated at $x^\dagger$, binary vectors are evaluated based on $f(c, x^\dagger)$. 
Because $x^\dagger$ is not optimal for $c \neq c^\dagger$, candidate solutions with $c = c^\dagger$ tend to be ranked as good solutions. 
Therefore, it tends to converge to $c^\dagger \neq c^\star$. 
The warm-starting strategy is expected to effectively prevent this issue.

\section{Hyper-Representation Technique}\label{sec:hyper-representation}

We address the difficulty of type-II interaction by introducing a parameterized map $\phi_w$ from a binary vector $c$ to a continuous vector $x$ with a parameter vector $w \in \R^{\ell}$. 
Arguably the simplest implementation of such a map is an affine map
\begin{equation}
\phi_w(c) = V c + b, \label{eq:phi-affine}
\end{equation}
where the parameter vector $w$ consists of the elements of $V \in \R^{n \times m}$ and the elements of $b \in \R^{n}$ and its dimension is $\ell = n (m + 1)$. 
We transform the original problem by introducing $\phi_w$ and try to minimize
\begin{equation}
    F(c, w) = f(c, \phi_w(c)).
\end{equation}
That is, instead of directly optimizing $(c, x)$, we optimize $c$ and $w$.

\paragraph{Why does it help?}
Let us first consider the ideal situation, where $\phi_w$ can represent the optimal map $\phi^\star$ in \eqref{eq:phistar} with a parameter $w^\star$, i.e., $\phi_{w^\star} = \phi^\star$.\footnote{It is not practical but indeed always possible to have a sufficiently high expressive ability. Because $\abs{\mathcal{C}} = 2^m$, we can prepare $2^m$ vectors $v_c \in \R^{n}$ and define $\phi_w(c) = v_c$ for each $c$, where $w \in \R^{2^m n}$ consists of all $v_c$.} 
Once such a map has been obtained, we can simply optimize $F(c, w^\star)$ for $c$ without further optimizing $w^\star$. The optimal solution to the original problem is $(c^\star, \phi_{w^\star}(c^\star))$, where $c^\star$ is the solution to $F(c, w^\star)$.
The point is that we have a single parameter vector $w^\star$ that is optimal over $c \in \mathcal{C}$. 
It is a significant difference from the original formulation, where the optimal continuous vector depends on $c$, which makes it difficult to correctly compare two $c$ and $c'$ under the corresponding optimal continuous vectors. 

It is more important to analyze whether we can obtain such an optimal $\phi_{w^\star}$ while $c$ is changing over time. Even if the representation ability is sufficiently high, it does not help to solve the original optimization problem if we can not obtain the optimal $w^\star$. 
For this purpose, we consider the synthetic problem described in \eqref{eq:f_II} with $\phi^\star(c) = V^\star c + b^\star$. For the parameterized map $\phi_w$, we employ the affine map defined in \eqref{eq:phi-affine}. Then, it is obvious that the optimal parameter is $V = V^\star$ and $b = b^\star$ and the representation ability is sufficient. 

The main question to answer here is whether the optimal parameter is the attractor of $F(c, x)$. 
To see this, we consider the derivatives of $F$ with respect to $w$. 
The partial derivative of $F$ with respect to $b_i$ is 
\begin{equation}
    \frac{\partial F}{\partial b_i}(c, w) = 2 (b_i - b_i^\star)
\end{equation}
and the partial derivative of $F$ with respect to $V_{i,j}$ is
\begin{equation}
    \frac{\partial F}{\partial V_{i,j}}(c, w) = 2 [ (V - V^\star) c c^\T ]_{i,j},
\end{equation}
where $[\cdot]_{i,j}$ represents the $(i,j)$-th element of an argument matrix.
We realize that $\frac{\partial F}{\partial b_i}(c, w)$ is independent of $c$. 
Therefore, $b$ is expected to approach $b^\star$ as long as a reasonable optimizer is used.
On the other hand, $\frac{\partial F}{\partial V_{i,j}}(c, w)$ depends on $c$. 
That is, the descent direction changes depending on the current $c$. 
However, if we consider to update $V$ following $- \nabla_V F(c, w)$, we have
\begin{align}
 V &\gets V - \epsilon \nabla_V F(c, w) \\
\implies  V - V^\star &\gets (V - V^\star) (I - 2 \epsilon c c^\T )
\end{align}
and $(I - 2 \epsilon c c^\T ) \preccurlyeq I$ for a sufficiently small $\epsilon$.
It implies that $V$ approaches $V^\star$ unless each element of $c$ is fixed to $0$. 
Therefore, $(V, b)$ approaches $(V^\star, b^\star)$ as long as they are updated in the descent direction and elements of $c$ are not fixed to $0$. 
This is the significant advantage of the proposed hyper-representation.

However, with the hyper-representation alone, the condition that elements of $c$ are not fixed to $0$ may be violated for a similar reason discussed in \Cref{sec:warm-start}. Therefore, it is important to employ the hyper-representation along with the warm-starting strategy. In the above situation, by employing the warm-starting technique, we can guarantee that elements of $c$ are not fixed to $0$ and $V$ converges to $V^\star$ as long as it is updated in the descent direction.\footnote{
The usefulness of the warm-starting employed with the hyper-representation can also be understood as follows.
Suppose that the objective function is arbitrary but the hyper-representation model is sufficiently expressive, i.e., there exists $w^\star$ such that $\phi_{w^\star} = \phi^\star$. 
Performing the warm starting is expected to converge $w$ to $w^\star$.
Then, the hyper-representation turns the original problem to $f(c, \phi^\star(c))$, which is a purely binary optimization problem, and no interaction between a continuous vector and a binary vector needs to be treated. In other words, as long as the hyper-representation model has sufficient expressive ability, one can decompose the mixed binary-continuous optimization problem into a continuous optimization and a binary optimization.
}



\section{IC\MakeLowercase{at}CMA}\label{sec:icatcma}
\begin{algorithm}[t]
\caption{ICatCMA} \label{alg:icatcma}
\begin{algorithmic}[1]
\REQUIRE $f: \mathcal{C} \times \mathcal{X}\to\R$, $\mu^{(0)}\in\R^\ell$, $\sigma^{(0)}\in\R_{>0}$, $C^{(0)}\in\R^{\ell\times\ell}$, $\phi_w: \mathcal{C}\to\R^n$, $q^{(0)}\in[0, 1]^{m}$
\STATE $p_\sigma^{(0)}=p_c^{(0)}=0, \delta^{(0)}=1, s^{(0)}=0, \gamma^{(0)}=0, t=0$
\FOR{$t=1, \dots, T_\mathrm{freeze}$}
    \STATE \textit{Sample} $c^{(t)} \sim p\left(c|q^{(0)}\right)$
    \STATE \textit{Sample} $w^{(t)}_k \sim p\left(w\Bigm|\mu^{(t)}, \left({\sigma^{(t)}}\right)^2 C^{(t)}\right)$ \textit{ for } $k=1, \dots, \lambda$
    \STATE \textit{Query} $\left\{f\left(c^{(t)}, \phi_{w^{(t)}_k}\left(c^{(t)}\right)\right)\right\}_{k=1}^\lambda$
    \STATE \textit{Update} $\mu^{(t)}, \sigma^{(t)}, C^{(t)}$ \textit{following CatCMA.}
\ENDFOR
\WHILE{termination conditions are not met}
    \FOR{$k=1, \dots, \lambda$}
        \STATE \textit{Sample} $c^{(t)}_k \sim p\left(c\bigm|q^{(t)}\right)$
        \STATE \textit{Sample} $w^{(t)}_k \sim p\left(w\Bigm|\mu^{(t)}, \left({\sigma^{(t)}}\right)^2 C^{(t)}\right)$
    \ENDFOR
    \STATE \textit{Query} $\left\{f\left(c^{(t)}_k, \phi_{w^{(t)}_k}\left(c^{(t)}_k\right)\right)\right\}_{k=1}^\lambda$
    \STATE \textit{Update} $\mu^{(t)}, \sigma^{(t)}, C^{(t)}, q^{(t)}$ \textit{following CatCMA.}
\ENDWHILE
\end{algorithmic}
\end{algorithm}

We describe the proposed method, \emph{ICatCMA}, which incorporates two proposed interaction treatment methods---the warm-starting strategy and the hyper-representation technique---into the original CatCMA.
\Cref{alg:icatcma} presents the algorithm of ICatCMA.
The continuous vector in CatCMA, $x\in\R^n$, is replaced with the parameter vector $w\in\R^\ell$ of the model $\phi_w$.
First, the warm-starting strategy kicks off (lines 2-7).
At each iteration, we sample a binary vector that is shared by $\lambda$ candidate solutions (line 2).
The parameters of the Gaussian distribution, $\mu, \sigma$, and $\Sigma$, are updated according to the CatCMA update while freezing the parameter vector for the Bernoulli distribution.
After $T_\mathrm{freeze}$ iterations, the standard CatCMA update is performed until termination conditions are met (lines 8-15).

\section{Experiments}\label{sec:experiments}

We conduct experiments to confirm the following hypotheses.
\begin{enumerate}
\item[RQ1] The original CatCMA fails to locate the optimal solution if the problem has type-I or II interaction. 
\item[RQ2] The proposed warm-starting (WS) strategy works effectively with CatCMA on problems with type-I interaction.
\item[RQ3] The proposed hyper-representation (HR) technique works effectively with CatCMA on problems with type-II interaction if employed with WS.
\item[RQ4] CatCMA with WS and HR, named CatCMA with interaction treatment (\emph{ICatCMA}), is also effective on problems with type-I and II interactions at the same time.
\item[RQ5] CatCMA is more efficient than ICatCMA when problems have no or only weak interaction.
\end{enumerate}

\subsection{Settings}

\paragraph{Problem}
We use the synthetic problems $f_{I}$ and $f_{II}$ (with $f_c(c) = \sum_{i=1}^{m} (1 - [c]_i)$) defined in \Cref{sec:prob}, as well as the following problem that combines type-I and type-II interactions:
\begin{equation}
    f_{III}(c, x) = \norm{x \odot c - \phi^\star(c)}^2 .
    \label{eq:f_III}
\end{equation}
For $f_{II}$ and $f_{III}$, we consider $\phi^\star(c) = \alpha V^\star c + b^\star$.
In addition, for $f_{II}$, we also test the case $\phi^\star(c)=\phi_\mathrm{tanh}^\star(c) := \tanh\left( \alpha V^\star c + b^\star \right)$.
$V^\star \in \R^{n \times m}$ and $b^\star \in \R^n$ are initialized randomly by a normal distribution, followed by the normalization such that $\norm{V^*}_F=\norm{b^*}_2=1$. 
They are independently generated for each problem instance, but for fair comparisons, the same instances (i.e., the same $V^*$ and $b^*$) are used for different algorithms.
With $\alpha = 0$, $f_{III}$ recovers $f_{I}$ where $b^\star$ is replaced with $\phi^\star(b^\star)$ and then we identify $f_{III}$ with $\alpha=0$ with $f_I$ in the following.
As mentioned in \Cref{sec:prob}, $\alpha \geq 0$ controls the strength of the interaction.
For $f_{II}$, we set $\alpha = 1, 2, 4, 8, 16$ and for $f_{III}$, we set $\alpha = 0, 1, 2, 4, 8, 16$.

 \paragraph{Algorithms}

 We compare four variants of CatCMA: the original CatCMA; CatCMA with the warm-starting strategy (WS-CatCMA); CatCMA with the hyper-representation technique (HR-CatCMA); CatCMA with both interaction treatments (ICatCMA).
For the hyper-representation technique, we use the linear model $\phi(c) = V c + b$, where $V\in\R^{n\times m}$ and $b\in\R^n$ are to be optimized by CatCMA.
For the warm-starting strategy, we set the number of freeze iterations as $T_\mathrm{freeze} = 5\times 10^2 \ell / \lambda$, where $\ell$ is the number of continuous variables to be optimized by CatCMA, which is $\ell = n(m + 1)$ if the hyper-representation is used; otherwise $\ell = n$. 
Our choice of $T_\mathrm{freeze}$ is based on the observation that CMA-ES usually requires function evaluations proportional to the number of continuous variables. 
Because $\ell$ differs for variants with and without HR, we also report the success rates for variants with $T_\mathrm{freeze} = 5000$ for fair comparison for some cases.
The initial mean vector $\mu^{(0)}$ of the normal distribution is set to the zero vector in $\R^{\ell}$ and the initial $\sigma^{(0)}$ is set to $1/(\ell+m)$.\footnote{
With this initialization, the expected values of the continuous vector with and without the hyper-representation for the initial distribution are both $0$. In this sense, we think that this initialization is fair for comparisons between algorithms with and without the hyper-representation.}
We follow \cite{hamano2024catcma} for the other initial parameters and hyper-parameters of CatCMA.

\paragraph{Experiment Procedure}
We run the four variants of CatCMA on 100 instances of each problem with varying parameters $\alpha, n, m$ described above.
We regard each run as successful if the function value smaller than $f_\mathrm{target} = 10^{-10}$ is found before the algorithm spends $10^6$ function evaluations.

\subsection{Results}




\Cref{tbl:fII} and \Cref{tbl:fIII} show the success rates of four algorithm variants on $f_{II}$ and $f_{III}$ with $\phi^\star$. Note that $f_{III}$ with $\alpha = 0$ is $f_{I}$. 
\Cref{tbl:fII_tanh} shows the success rates of four algorithm variants on $f_{II}$ with $\phi_\mathrm{tanh}^\star$. 

\paragraph{Effect of Warm-Starting}
The effect of the warm-starting strategy is most pronounced on $f_{I}$ ($f_{III}$ with $\alpha = 0$), a problem with type-I interaction. 
CatCMA variants without the warm-starting strategy frequently fail to locate the optimal binary vector as shown in \Cref{tbl:fIII}.
On the other hand, the variants with the warm-starting strategy tend to locate the optimal solution with a higher success rate.
In particular, by comparing the results with $T_\mathrm{freeze} = 5 \times 10^2 \ell / \lambda$ and $T_\mathrm{freeze} = 5000$, we notice that a higher $T_\mathrm{freeze}$ contributes to achieving a higher success rate (note that $5 \times 10^2 \ell / \lambda < 5000$). 
As shown in \Cref{tbl:fIII}, on $f_I$, the success rate significantly improves with WS.
In particular, when solely employing WS, the success rate reaches almost 1.0 for all cases of $f_I$.
Also in $f_{II}$ and $f_{III}$, the success rates when employing WS are almost equal or greater than ones when not employing WS as shown in \Cref{tbl:fII} and \Cref{tbl:fIII}.

\paragraph{Effect of Hyper-Representation}
The effectiveness of the hyper-representation technique is revealed in \Cref{tbl:fII,tbl:fIII} with $\alpha > 0$. 
We see that the hyper-representation alone is insufficient for treating the difficulty of type-II interaction by comparing the results of CatCMA variants without the warm-starting strategies on \Cref{tbl:fII}. 
All four CatCMA variants successfully locate the global optimum on $f_{II}$ with $\alpha \leq 2$, whereas the success rates drop significantly with increasing $\alpha$ except for the variant with both the warm-starting strategy and the hyper-representation technique. 
The warm-starting strategy is slightly advantageous on $f_{II}$ but the effect of the warm-starting strategy alone on the difficulty of type-II interaction is not sufficient. 
The combination of the warm-starting strategy and the hyper-representation technique improves the success rate significantly as expected. 

\begin{table}[t]
\centering
\caption{Results on $f_{II}$ ($m = n = 5$). }\label{tbl:fII}
\begin{subtable}{\hsize}
\centering
\caption{$T_\mathrm{freeze} = 5\times 10^2 \ell / \lambda$.}\label{tbl:fII_adaptive}
\begin{tabular}{c|ccccc}
\toprule
algorithm & $\alpha = 1$ & $2$ & $4$ & $8$ & $16$ \\
\midrule
CatCMA & 1.0 & 0.99 & 0.46 & 0.01 & 0.0 \\
WS-CatCMA & 1.0 & 0.99 & 0.65 & 0.21 & 0.1 \\
HR-CatCMA  & 1.0 & 1.0 & 0.54 & 0.0 & 0.0 \\
ICatCMA  & 1.0 & 1.0 & 1.0 & 1.0 & 0.96 \\
\bottomrule
\end{tabular}
\end{subtable}%
\\
\begin{subtable}{\hsize}
\centering
\caption{$T_\mathrm{freeze} = 5000$.}\label{tbl:fII_fixed}
\begin{tabular}{c|ccccc}
\toprule
algorithm & $\alpha = 1$ & $2$ & $4$ & $8$ & $16$ \\
\midrule
WS-CatCMA & 1.0 & 0.96 & 0.49 & 0.2 & 0.07 \\
ICatCMA  & 1.0 & 1.0 & 1.0 & 1.0 & 1.0 \\
\bottomrule
\end{tabular}
\end{subtable}
\end{table}

\paragraph{Representation ability of $\phi_w$}
We observe the success rate on $f_{II}$ with $\phi^\star=\phi_\mathrm{tanh}^\star$ (\Cref{tbl:fII_tanh}) to investigate the case that the representation ability of $\phi_w$ is not completely sufficient (as $\phi_w(c)$ is the linear function $V c + b$).
It is observed that ICatCMA is still effective when the representation ability of $\phi_w$ is not completely sufficient. For all $\alpha$, the success rates improve with HR in \Cref{tbl:fII_tanh}.
Note that with $\phi_\mathrm{tanh}^\star$, the problem seems to become easier to solve since the impact of $\alpha$ becomes weaker.

\begin{table}[t]
\centering
\caption{Results on $f_{III}$ ($m = n = 5$). }\label{tbl:fIII}
\begin{subtable}{\hsize}
\centering
\caption{$T_\mathrm{freeze} = 5\times 10^2 \ell / \lambda$.}\label{tbl:fIII_adaptive}
\begin{tabular}{c|cccccc}
\toprule
algorithm & $\alpha = 0$ & $1$ & $2$ & $4$ & $8$ & $16$ \\
\midrule
CatCMA & 0.15 &  0.03 & 0.03 & 0.01 & 0.01 & 0.0 \\
WS-CatCMA & 0.99  &  0.16 & 0.02 & 0.01 & 0.0 & 0.01 \\
HR-CatCMA & 0.14 &  0.01 & 0.0 & 0.01 & 0.0 & 0.0 \\
ICatCMA & 0.56 &  0.52 & 0.19 & 0.12 & 0.09 & 0.06 \\
\bottomrule
\end{tabular}
\end{subtable}%
\\
\begin{subtable}{\hsize}
\centering
\caption{$T_\mathrm{freeze} = 5000$.}\label{tbl:fIII_fixed}
\begin{tabular}{c|cccccc}
\toprule
algorithm & $\alpha = 0$ & $1$ & $2$ & $4$ & $8$ & $16$ \\
\midrule
WS-CatCMA & 1.0  &  0.15 & 0.0 & 0.0 & 0.0 & 0.02 \\
ICatCMA  & 0.85 &  0.81 & 0.51 & 0.53 & 0.32 & 0.22 \\
\bottomrule
\end{tabular}
\end{subtable}
\end{table}

%
%
%

\begin{table}[t]
\centering
\caption{Results on $f_{II}$ with $\phi_\mathrm{tanh}^\star$ ($m = n = 5$). $T_\mathrm{freeze} = 5\times 10^2 \ell / \lambda$.}\label{tbl:fII_tanh}
\begin{tabular}{c|ccccc}
\toprule
algorithm & $\alpha = 1$ & $2$ & $4$ & $8$ & $16$ \\
\midrule
WS-CatCMA & 1.0 & 1.0 & 0.97 & 0.94 & 0.85 \\
ICatCMA & 1.0 & 1.0 & 1.0 & 1.0 & 0.96 \\
\bottomrule
\end{tabular}
\end{table}

%


\paragraph{Combined Difficulties of Interaction Types}
\begin{figure}[t]
    \centering
    \includegraphics[width=1\linewidth,clip,trim=10 10 10 10]{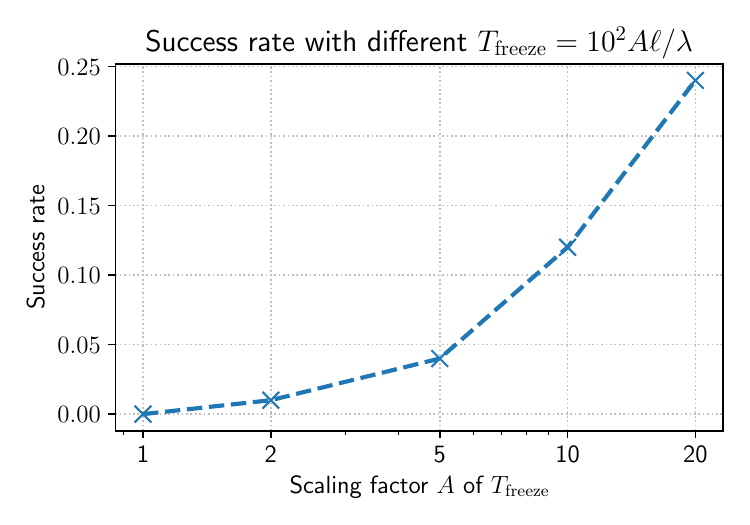}
    \caption{Success rates (over $100$ trials) of ICatCMA when $T_\mathrm{freeze}= A\cdot10^2\ell/\lambda$ with $A=[1, 2, 5, 10, 20]$ on $f_{III}$ with $\alpha=16$. 
    }
    \label{fig:different_T_freeze}
\end{figure}
As we can see in \Cref{tbl:fIII}, the combination of the difficulties of interaction type-I and interaction type-II makes the problem hard to solve for all variants. 
The success rates of the CatCMA variants drop significantly from $f_{II}$ to $f_{III}$ (by introducing the difficulty of interaction type-I into $f_{II}$) with the same $\alpha$.
Moreover, they drop significantly from $\alpha = 0$ (without interaction type-II) to $\alpha = 1$ (by introducing the difficulty of interaction type-II into $f_{I}$) except for ICatCMA.
ICatCMA improves the success rate significantly on $f_{III}$. 
Still, the success rate decreases as the strength of the type-II interaction, i.e., $\alpha$, increases (the bottom rows in \Cref{tbl:fIII_adaptive} and \Cref{tbl:fIII_fixed}). 
We conduct additional experiments to explore why the success rate drops significantly when using ICatCMA on $f_{III}$ with large $\alpha$. In Figure~\ref{fig:different_T_freeze}, we observe the success rates associated with different $T_{\text{freeze}}$ values, $A \cdot 10^2\ell/\lambda$ with $A = [1, 2, 5, 10, 20]$, for $f_{III}$ with $\alpha = 16$. It is noted that the success rates increase as $T_{\text{freeze}}$ is incremented, indicating that ICatCMA could solve this problem with greater certainty if $T_{\text{freeze}}$ is adequately large. Investigating more suitable settings for $T_{\text{freeze}}$ remains an ongoing challenge.

\paragraph{Efficiency}
\begin{figure}[t]
    \centering
    \includegraphics[width=1\linewidth,clip,trim=5 10 5 5]{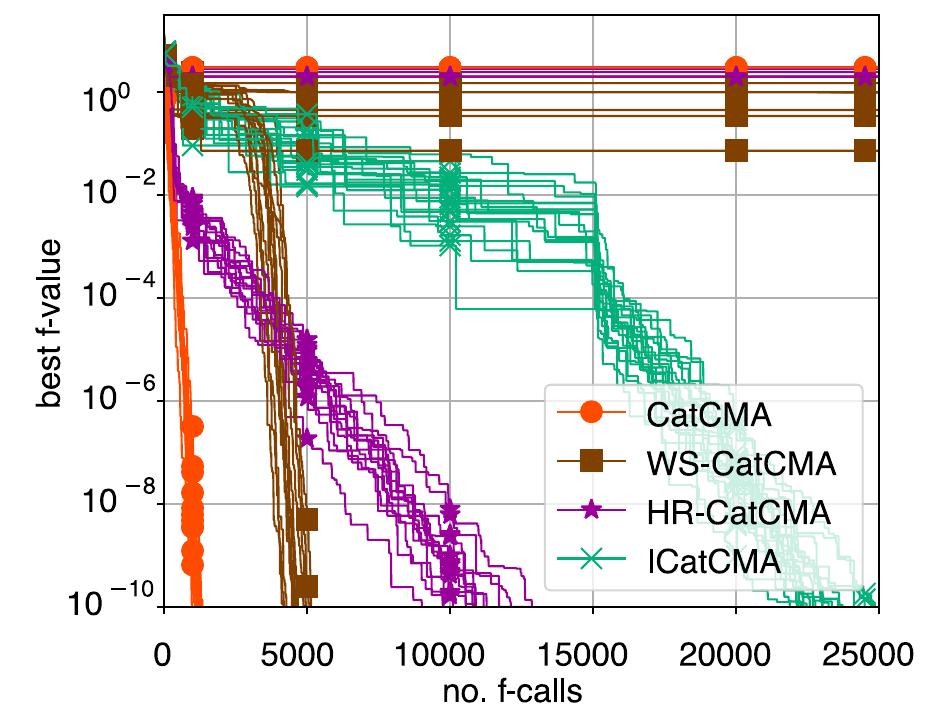}
    \caption{The best-so-far $f$-value evaluated on $f_{II}$ with $\alpha=4$. Each optimizer runs on $20$ problem instances.}
    \label{fig:efficiency}
\end{figure}
While warm-starting has proven effective for the interactions considered in the previous experiment, it is posited that if the optimization problem is sufficiently simple for the standard CatCMA, $T_{\text{freeze}}$ may simply add unnecessary overhead to the optimization time.
In addition, the hyper-representation method also alleviates the difficulties caused by interaction type-II by introducing high redundancy into the continuous variables, and then the dimension of the continuous variables significantly increases ($n$ to $\ell = n(m+1)$ in the above experimental setting).
Clearly, the slowdown in optimization due to increased dimensionality is unavoidable and may be troublesome.
We conduct additional experiments to observe the effect of dimensionality.
\Cref{fig:efficiency} shows the best-so-far $f$-value evaluated on $f_{II}$ with $\alpha=4$, which can be solved with the standard CatCMA with a probability of about 50\% (as shown in \Cref{tbl:fII}).
CatCMA requires the fewest $f$-calls to reach the target $f$ values of $10^{-10}$ or below when it successfully finds the optimum. It locates the near-optimal solutions in around $100$ iterations.
WS-CatCMA exhibits a higher success rate on this problem, but it requires more $f$-calls. 
The convergence rate is more or less the same as that of CatCMA, while it spends around $3000$ $f$-calls before converging, which corresponds to $T_\mathrm{freeze}$ iterations. It demonstrates that the warm-starting strategy seems to be an unnecessary overhead in this problem.
The convergence rates of HR-CatCMA and ICatCMA are similar, and they are significantly slower than those of CatCMA and WS-CatCMA. 
It is due to the increase of the dimension of the continuous vector. 
Thus, in problems such as $f_I$ where HR is not always necessary, the hyper-representation technique slows down the convergence, leading to inefficient optimization. 
ICatCMA suffers from both the overhead due to $T_\mathrm{freeze}$ and the slowdown due to the increased dimensionality, though the success rate is the highest among these four variants.

%
%
%
%

\section{Conclusion}\label{sec:conclusion}

In this study, we study two types of difficulty in mixed binary-continuous black-box optimization problems, that is, type-I interaction and type-II interaction.
In the problem with type-I interaction (exampled as $f_I$ in \Cref{eq:f_I}), the continuous variables are masked by the binary variables, and then a portion of the values of the continuous variables become completely meaningless for certain binary ones. 
In the problem with interaction type-II (exampled as $f_{II}$ in \Cref{eq:f_II}), the values of the continuous ones which minimize the objective function are determined by the binary ones and then it becomes very strenuous to optimize binary ones and continuous ones separately.
We propose two techniques for handling such interactions: warm-starting strategy and hyper-representation technique.
CatCMA with these two techniques, named ICatCMA, is tested on test problems. The findings are as follows.
\begin{itemize}
\item CatCMA fails to locate the optimal solution if the problem has type-I or II interaction. 
\item The warm-starting strategy works effectively on problems with type-I interaction.
\item The hyper-representation technique alone is not sufficient to tackle type-II interaction but works effectively with the warm-starting strategy.
\item ICatCMA is effective on problems with type-I and II interactions at the same time.
\end{itemize}
Meanwhile, we observe the inefficiency of ICatCMA compared to the original CatCMA when solving problems that CatCMA can easily solve.

Our findings contribute to understanding the interaction between binary and continuous variables, which we believe is common in real-world applications. However, our work has certain limitations. We conclude this paper with a discussion of potential areas for improvement.
\paragraph{Efficiency of the proposed method}
The proposed methods involve two factors that impact the efficiency of optimization: the freezing time $T_{\mathrm{freeze}}$ and the selection of the model $\phi_w$. 
While the warm-starting technique always succeeds if $T_{\mathrm{freeze}}$ and the representational ability of $\phi_w$ are sufficiently high, the speed of optimization decreases with a larger $T_{\mathrm{freeze}}$. 
The selection of $\phi_w$ also presents practical issues; although a redundant $\phi_w$ can accurately represent $\phi^\star$, it also leads to slower convergence by increasing the dimensionality of the continuous variable $w$. 
We have observed these trade-off problems in \Cref{fig:efficiency}.
Balancing the success rate and convergence speed is one of the most important remaining tasks.

\paragraph{Theoretical understanding of the difficulties}
In continuous optimization, it is well-known that the difficulty of interactions among variables in the objective function $f$ can often be quantified by the condition number of $f$'s Hessian. 
However, in mixed categorical-continuous black-box optimization, there is no consensus on how to quantify the strength of these interactions. 
We characterize the difficulties of interaction types I and II through specific functions, $f_I$, $f_{II}$, and $f_{III}$, and a specific parameter, $\alpha$. 
Developing a general perspective on the interactions in such problems remains an important area for future research.

\paragraph{Comprehensive empirical study}
It cannot be said that the validation in this paper was conducted on a wide range of mixed categorical-continuous problems.
First, we limit the focus of this study to mixed binary-continuous optimization for simplicity, although the proposed approach can be easily extended to mixed categorical-continuous optimization. Empirical evaluation on mixed categorical-continuous problems remains an important direction for future work.
Moreover, the tested combinations of continuous and binary variable dimensions are limited in this paper.
As discussed in \Cref{sec:experiments}, increasing the dimensionality of continuous variables may introduce challenges in certain scenarios, requiring further investigation.
Finally, evaluating the proposed approach on real-world applications, such as model merging \cite{akiba2024evolutionary}, would help demonstrate its practical usefulness. These are all important directions for future research.



\end{document}